\documentclass[runningheads]{llncs}
\usepackage{graphicx}
\usepackage{comment}
\usepackage{amsmath,amssymb} 
\usepackage{color}
\usepackage{multirow}
\usepackage[hidelinks]{hyperref}

\usepackage[width=122mm,left=12mm,paperwidth=146mm,height=193mm,top=12mm,paperheight=217mm]{geometry}
\usepackage{pdfpages}

\begin{document}
\pagestyle{headings}
\mainmatter
\def\ECCVSubNumber{2989}  

\title{JNR: Joint-based Neural Rig Representation\\for Compact 3D Face Modeling} 

\titlerunning{JNR: Joint-based Neural Rig Representation}
\authorrunning{N. Vesdapunt et al.}
\author{Noranart Vesdapunt \and Mitch Rundle \and HsiangTao Wu \and Baoyuan Wang}
\institute{Microsoft Cloud and AI \\
\email{\{noves, mitchr, musclewu, baoyuanw\}@microsoft.com}}

\maketitle
\setlength{\belowcaptionskip}{-5pt}

\begin{figure*}[t]
\centering
\includegraphics[width=1.0\textwidth]{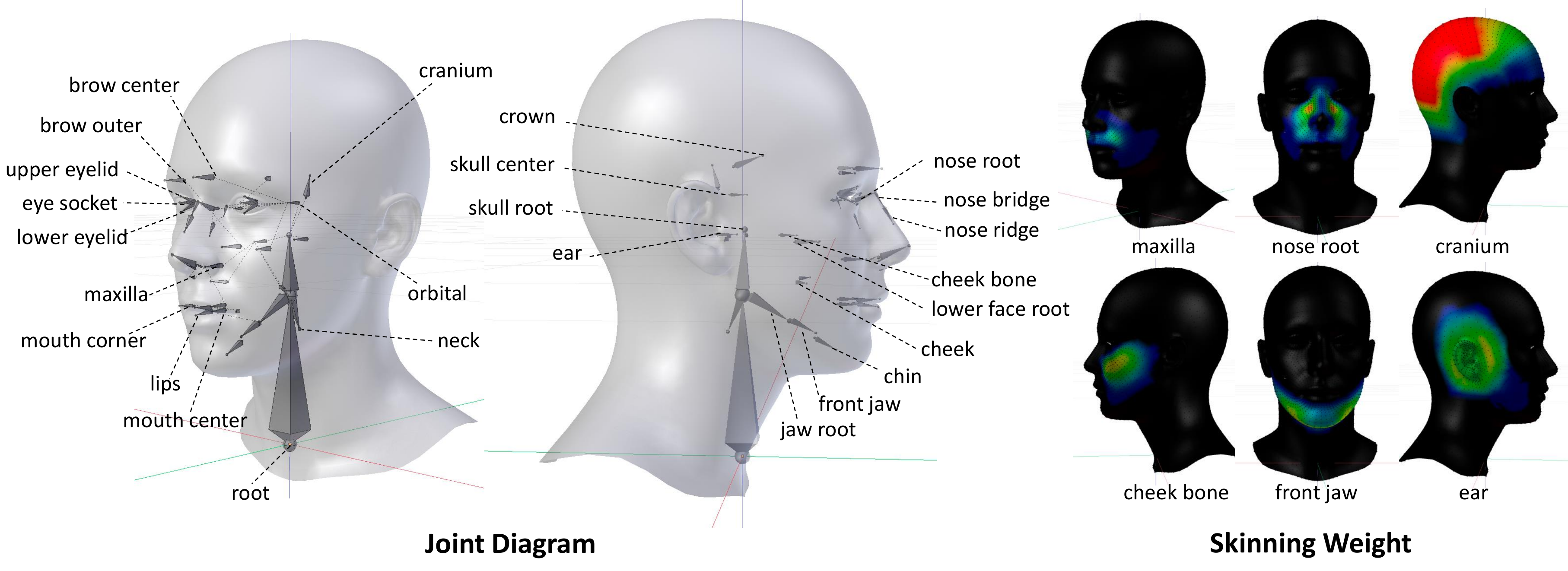}
\caption{An illustration of 52 joints in our joint-based model with hand-painted skinning weight. Our joints are defined hierarchically and semantically by following human anatomy without any training data. Our model is very compact, i.e., only a single mesh (5k vertices) with 9k floating points for skinning weight.}
\label{fig:joint_diagram}
\end{figure*}

\begin{abstract}
In this paper, we introduce a novel approach to learn a 3D face model using a joint-based face rig and a neural skinning network. Thanks to the joint-based representation, our model enjoys some significant advantages over prior blendshape-based models. First, it is very compact such that we are orders of magnitude smaller while still keeping strong modeling capacity. Second, because each joint has its semantic meaning, interactive facial geometry editing is made easier and more intuitive. Third, through skinning, our model supports adding mouth interior and eyes, as well as accessories (hair, eye glasses, etc.) in a simpler, more accurate and principled way. We argue that because the human face is highly structured and topologically consistent, it does not need to be learned entirely from data.  Instead we can leverage prior knowledge in the form of a human-designed 3D face rig to reduce the data dependency, and learn a compact yet strong face model from only a small dataset (less than one hundred 3D scans). To further improve the modeling capacity, we train a skinning weight generator through adversarial learning. Experiments on fitting high-quality 3D scans (both neutral and expressive), noisy depth images, and RGB images demonstrate that its modeling capacity is on-par with state-of-the-art face models, such as FLAME and Facewarehouse, even though the model is 10 to 20 times smaller.  This suggests broad value in both graphics and vision applications on mobile and edge devices.

\keywords{Face Modeling, 3D Face Reconstruction, GANs}

\end{abstract}

\begin{figure*}[t]
\centering
\includegraphics[width=1.0\linewidth]{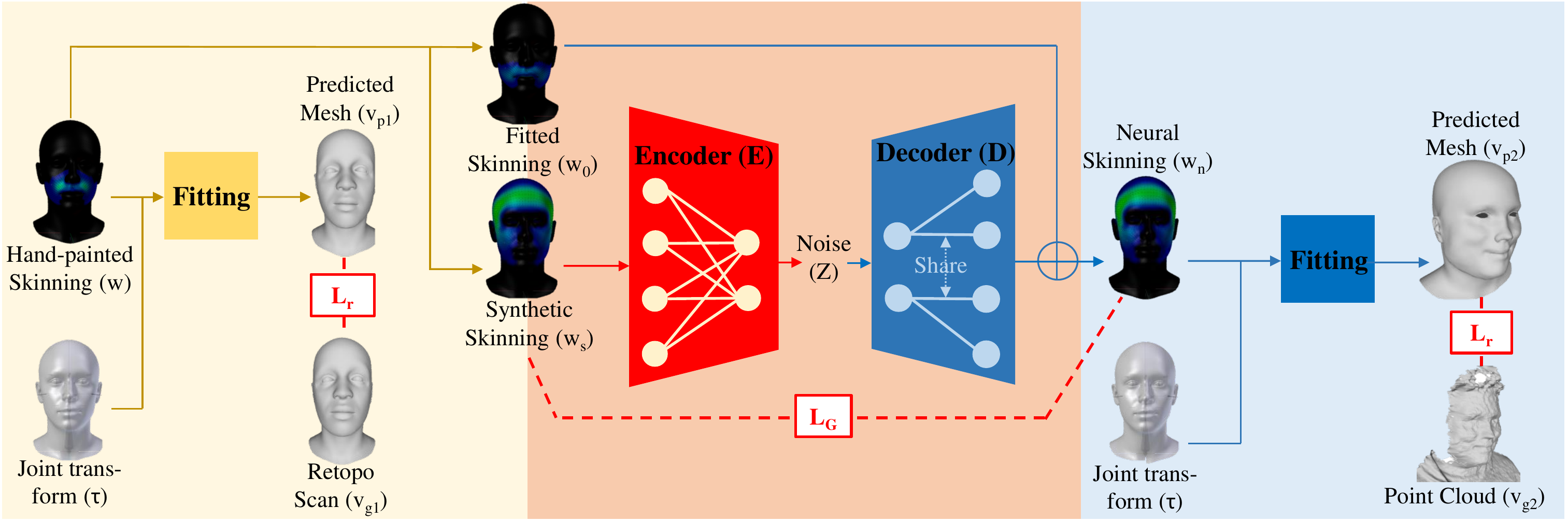}
\caption{Our pipeline has 3 stages. (1) Optimize joint transformation ($\tau$) and hand-painted skinning weight on retopologized scans to generate learned linear skinning weight, and synthetic skinning weight training data. (2) Train an autoencoder to output neural skinning weight then finetune the decoder to cover entire $Z$ distribution. (3) Fix the decoder then optimize $\tau$ and $Z$ on point cloud with the same loss as (1), but correspondences are built by ICP. Note that only (3) is needed at test time.}
\label{fig:pipeline}
\end{figure*}

\section{Introduction}

Parametric face models are popular representations for 3D face reconstruction and face tracking, where the face geometry is represented in a lower dimensional space than raw vertices. 3D Morphable Face Model (3DMM) proposed in 1999 \cite{3DMM} is still the dominating representation, although there are a few more recent models such as BFM \cite{BFM2009} and FLAME \cite{FLAME} which showed stronger modeling capacity with more training data. They all advocate learning the face model from a large number of high quality 3D scans. Collecting a large scan dataset, however, comes with a high cost along with the need for complicated and labor-intensive post processing such as cleanup, retopologizing and labeling to ensure accurate point to point correspondence before learning the face model. The previous state-of-the-art FLAME \cite{FLAME} model was learned from 33k registered scans which is obviously not easy to scale. 

Model size is another important aspect to democratizing 3D face applications, which unfortunately has not received much attention. A cheap, efficient solution is required to support low memory 3D face reconstruction and tracking, especially on edge devices. Even on the cloud, efficiency is compelling as a means of saving cost. The highest capacity model from FLAME was shown with 300 identity basis which is over 4 million floating points, and while BFM \cite{BFM2009} has fewer identity basis, the larger number of vertices increases the size to 32 million floating points. The challenging issues of collecting data and reducing model size lead us to explore a strong prior that can reduce the need for both of them. 

A few recent works \cite{FML,nonlinear3dmm} propose to learn a face model from massive wild images or videos. Although impressive results were shown, those models were primarily designed for RGB input. Moreover, the face model is implicitly embedded within the trained network, so it is unclear how it can be extracted to serve other purposes, such as fitting to depth data and interactive editing (such as ZEPETO \cite{zepeto} App). With the ubiquity of depth cameras (e.g., Microsoft Azure Kinect), fitting 3D face models to depth or point cloud data may be an increasingly important problem for both face reconstruction and tracking \cite{Weise:2011:RPF}.

In this paper, we propose to use a new face representation that leverages a human designed, joint-based, parametric 3D face rig with learned skinning weights to model the shape variations. At the heart of our model, we bring human prior knowledge into the modeling which significantly reduces the data that would otherwise be needed to learn everything from scratch. Although joint-based representations receive little attention in literature, they are standard in the gaming and movie industries. To build our model, we used a standard low-poly base mesh whose topology supports anatomically plausible deformations with minimal artifacts. We created a deformation skeleton and skinned the mesh to it using hand-painted skinning weights (see Fig.\ref{fig:joint_diagram}). Our model can be replicated easily using the details provided in supplementary materials, and it needs to be created only once. It can then be used in applications just like any other face models (FLAME \cite{FLAME}, BFM \cite{BFM2009}, Facewarehouse \cite{facewarehouse}, etc). Our technique is also compatible with joint-based face models publicly available on the internet. The model capacity is defined by the number of joints, their bind poses and the skinning weights, which are typically fixed. The skinning weights, however, do not need to be fixed and are a key to further increasing the representation capacity.

To learn skinning weights, we collected 94 high quality scans and retopologized them to our face topology. We took 85 of our scans as training data, then fit one fixed global skinning weight matrix across all the scans through an iterative optimizer.  This shows noticeable improvement over the hand-painted skinning weights, and with this model we achieve lower error than FLAME 49 while being 30 times smaller. We then further increased our model capacity by training a neural network to encode the skinning weight space into 50 parameters. We test the effectiveness of the neural skinning weights on BU-3DFE neutral and we can approximate FLAME 300 scan-to-mesh error, while staying 20 times more compact. Our model also handles expressions out of the box, but we added a subset of expression blendshapes from \cite{MultiFaceRetarget19} to further increase capacity. Our final model is on-par with FLAME 300 while remaining 10 times more compact (evaluate on expressive scans from BU-3DFE). Our fitted face mesh enjoys the benefit of joint model as it facilitates accurate placement of mouth interior, eyes and accessories.  Furthermore, the semantically meaningful parameters facilitate post-fit editing. We summarize our main contribution as following:
\begin{enumerate}
\itemsep0em 
\item We propose a new 3D face representation based on a joint-based face rig model that incorporates strong human prior.  This allow us to reduce the required number of training scans, reduce the size of the model, accurately place mouth interior, eyes and hair, as well as support editing.

\item We propose a neural skinning weight generator to significantly increase our joint-based model capacity. To preserve the compactness of our model, we design a group-wise fully connected layer to reduce the model size 51 times.

\item We tested our model on BU-3DFE \cite{bu3dfe}, retopologized scans, Azure Kinect depth dataset, 2D images with 2D landmarks, and retargeting. Results show that our model can achieve similar error with FLAME 300 \cite{FLAME} with 20 times smaller for neutral scans, and 10 times smaller for expressive scans.
\end{enumerate}

\section{Related Work}

\textbf{Linear 3D Face Modeling} In 1999, Blanz and Vetter \cite{3DMM} are the first to introduce morphable model by learning PCA from 200 scans. Ten years later, Basel Face Model (BFM) \cite{BFM2009} was built on another 200 scans using a nonrigid extension of iterative closest point algorithm (NICP) \cite{NICP}. Recently,  Large Scale Facial Model (LSFM) \cite{LSFM} built a large scale version of BFM by learning from 10k scans. Wu et al. \cite{WU} added a joint for jaw rotation to significantly reduce the number of required blendshapes. Inspire by \cite{WU}, FLAME \cite{FLAME}, added joints to allow eye, jaw and head rotation. FLAME was trained with 33k scans, and is considered to be the current state-of-the-art. While the trend is towards larger amounts of training data for building face models, we believe that it is hard to democratize such methods due to the high cost of data collection as well as privacy concerns. The authors of \cite{selfsupervised_ayush} propose to learn a geometry and texture correction in a PCA space based of 3DMM. JALI \cite{JALI} proposes a face model specifically for speaking movement. Combined Face \& Head Model (CFHM) \cite{CFHM} designed a combination of Liverpool-York Head Model (LYHM) \cite{LYHM} and LSFM \cite{LSFM} to achieve a face model that can perform well on both head and face region. York Ear Model \cite{ear} proposes an ear modeling. Muscle-based model \cite{muscle} seeks to explain human movement anatomically similar to us, but express in blendshapes and jaw joint. We define our model anatomically purely in joints and introduce neural skinning weight to increase model capacity.   \\

\noindent\textbf{Nonlinear 3D Face Modeling}
Fernandez et al. \cite{fernandez} designs a convolutional encoder to decouple latent representation and apply a multilinear model as a decoder to output depth images. Non-linear 3DMM \cite{nonlinear3dmm} proposes to learn face model by convolutional neural network on 300W-LP \cite{300w} with RGB input and the outputs are position map and texture in UV space. Face Model Learning from Videos (FML) \cite{FML} learns face model on fully-connected layers to output shape and texture in the vertex space instead of UV space, and is designed to tackle the lack of training data through unsupervised learning on a large-scale video dataset (VoxCeleb2 \cite{voxceleb2}). FML is the closest work to ours in term of addressing limited 3D scan datasets, but it was not designed for point cloud fitting.    

While autoencoder family \cite{COMA,Jiang_2019_CVPR,Dai_2019_ICCV,doi:10.1111/cgf.13830} achieves impressive error on retopologized mesh, which we call known-correspondence, it is unclear how to accommodate other tasks. We consider FLAME as a more complete face model as it can use NICP to fit unknown correspondence, and the authors demonstrate 2D fitting and retargeting in their paper. To make a fair comparison, we retrained COMA on our 85 training scans, but COMA \cite{COMA} does not generalize to BU-3DFE at all. COMA \cite{COMA} and Li et al. \cite{doi:10.1111/cgf.13830} require 20k scans to achieve their results. The model size is also larger, e.g., Dai et al. \cite{Dai_2019_ICCV}’s decoder alone is 17.8M Floats and Jiang et al. \cite{Jiang_2019_CVPR}’s encoder alone is 14.0M Floats, while our entire model is only 0.2M Floats. \\

\noindent\textbf{Personalized 3D Face Modeling}
High quality 3D face modeling is already possible if the model is for a specific person. Past works \cite{incrementalfacetracking,mesoscopicgeometry} capture images from multiple views and apply multi-view stereo reconstruction to obtain 3D face mesh. Using videos instead of multiview \cite{pablo_geomfromvideo}, \cite{GZCVVPT16} is also popular for reducing the cost of camera rig. Using only internet photos is also possible in \cite{wildphotobasedreconstruction} where the authors personalize normal maps to obtain user's mesh. Our work indirectly creates personalized 3D face models using the neural skinning weights. \\

\noindent\textbf{Skinning Weight Learning}
Bailey et al. \cite{bailey2018fast} and Neuroskinning \cite{liu2019neuroskinning} propose to learn skinning weight to replace manual effort in painting the skinning weight for character animation. Although the idea of using neural network in Neuroskinning is similar to ours, the goal is different as we wish to increase model capacity for face modeling. Moreover, Neuroskinning takes a 3D model as an input, while we train our network as a GAN. Perhaps a direct comparison would be inputting both 3D scan and joint transformation into our network, but the network will not generalize well due to the small amount of scans in training set. The network size is also directly counted towards face model size, so Neuroskinning's network design would defeat the purpose of compactness of our joint model.  \\

\section{Methodology}
We first define the plain joint-based model with linear blend skinning where the weights are fixed, then introduce an advanced version where we use a network to generate skinning weights adaptively though optimizing the latent code in section \ref{sec:neuralskinning}. Fig.\ref{fig:pipeline} illustrates the overview of our system pipeline.
\subsection{Joint-based Rigging Model}
\subsubsection{Joint-based Face Representation}
We use vertex based LBS similar to FLAME \cite{FLAME} with $N$ = 5236 vertices, but $K$ = 52 joints. This can be viewed as replacing the blendshapes in FLAME with 47 more joints which are defined as:

\begin{equation}
M_k = B^{-1} * \tau * M_{p}
\end{equation}

Where $B$ is a fixed 4x4 bind pose matrix and the fitting variables are rigid transformation $\tau$, which includes Euler rotation ($R$), translation ($T$) and scaling ($S$). Our joint are defined hierarchically, where the global joint transformation $M_k$ is defined by recursive multiplications from the parent joint $M_{p}$, to the root joint $M_{p0}$ (an identity matrix). We first define the root joint to allow global pose, and design all the other joints hierarchically such that each level covers a smaller area of the mesh and therefore finer detail. We then evaluate and visualize the fitting error on high quality scans and adjust the current joints and/or add smaller joints, and repeat the process until we reach desired level of morphing detail. As we only allow specific transformation on each joint, and we enforce bilateral symmetry in most cases, we end up with only 105 variables as opposed to 468 (9 DOF x 52 joints). The full details can be found in supplementary.

\subsubsection{Linear Blend Skinning}
\label{sec:skin}
We apply $M_k$ and LBS on each vertex.

\begin{equation}
v' = \sum_{k=1}^{K} (w_k * v * M_k)
\end{equation}

Skinning weights ($w$) are defined in a 2-dimensional $N * K$ matrix, and $v$ is the vertex position. We first initialize $w$ with skinning weights created manually using the weight painting tool in Blender. Even with some expertise in the area, it is difficult to produce good results in regions influenced by many joints, and quite intractable to find weights that work well across a wide range of identities. Therefore, we took a data-driven approach and regressed $w$ over a set of ground-truth meshes, rather than tweaking them manually to fit all kinds of scans. As we design our joint to mainly model the local geometry variations, more than half of the joints only affect small regions of the face. While Dictionary 3DMM \cite{Ferrari2015} uses sparse PCA to enforce local sparsity, our skinning weights $w$ are naturally very sparse. Similar to symmetry constraint in 3DMM \cite{Ankur2011}, our joints are also bilaterally symmetrical, so too are the skinning weights. Sparsity and symmetry reduce the number of variables from $N * K = 272,272$ to 8,990 floating points (30.3x). Fig.\ref{fig:joint_diagram} shows a sample of our skinning weights. As we only need 9k floats to represent skinning weights, we optimize them as unknown parameters and fit them across all the scans in the training set. As a result, the learned skinning weights do not produce sharp artifacts that would be difficult to fix manually.

\subsubsection{Losses}
In order to eliminate the correspondences error, we use the commercial tool Wrap3 \cite{wrap3} to manually register all the scans into our topology. Once we establish the dense point-point correspondences across all the scans and our model, the fitting is conducted by minimizing RMSE:

\begin{equation}
L_{v} = \frac{1}{N}\sqrt{\sum_{n=1}^{N} (v_{p,n} - v_{g,n})^2}
\end{equation}

Where $v_p$ is the predicted vertex and $v_g$ is the ground-truth vertex. When fitting to depth test data, we use NICP to build the correspondences between predicted mesh and point cloud. Similar to mean face regularization, we initialize local transforms to identity transformation ($I$) and add a diminishing regularization term to limit joint transformations while NICP converges.

\begin{equation}
L_{m} = \frac{1}{K}\sum_{k=1}^{K} (|R_k| + |T_k| + |S_k-1|)
\end{equation}

We manually limit the transformations to keep deformations within anatomically plausible bounds. This is similar to PCA coefficient limit, as FLAME also generates artifacts if the coefficients exceed [-2, 2].
\begin{equation}
L_{x} = \frac{1}{K}\sum_{k=1}^{K} \begin{cases}
      |x_k-x_{max}|, & \text{if}\ x_k > x_{max} \\
      |x_k-x_{min}|, & \text{if}\ x_k < x_{min} \\
      0, & \text{if}\ x_{min} <= x_k <= x_{max} \\
    \end{cases}
\end{equation}

Where $x$ is a substituted annotation for $R$, $S$, $T$, and we have $L_{x}$ on all of them. We also follow the standard regularization in \cite{facewarehouse,pablo_geomfromvideo,wildphotobasedreconstruction,DBLP:conf/cvpr/TewariZ0BKPT18,incrementalfacetracking}.

\begin{equation}
L_{p} = \frac{1}{N}\sqrt{\sum_{n=1}^{N} (\nabla^2(v_{p,n}) - \nabla^2(v_{g,n}))^2}
\end{equation}

Where $\nabla^2$ is a Laplacian operator between $v_p$ and $v_g$ to ensures the overall topology does not deviate too much. Our final loss is the summation of all losses.

\begin{equation}
L_r = L_{v} + \lambda_m L_{m} + \lambda_x (L_{R} + L_{T} + L_{S}) + \lambda_p L_{p} 
\label{loss}
\end{equation}

We empirically set the weight for each loss with $\lambda_m$ = 0.03, $\lambda_x$ = $\lambda_p$ = 0.3.

\subsection{Neural Skinning Weight}\label{sec:neuralskinning}
\subsubsection{Model Capacity}
As we seek to further increase the capacity of our joint-based model, we could consider applying free form deformation or corrective blendshapes on top of the fitting result, but that would nullify the ability afforded by the joint-based rig to accessorize the mesh. Although a local region expression blendshapes is tolerable, identity blendshapes that effect the mesh globally will also break the accessorizing. We could also add more joints, either manually or through learning, but this begins to dilute the semantic meaning of the joints, and furthermore can lead to deformation artifacts caused by too many joints in close proximity. Instead, we chose to fit the skinning weights to each test subject thereby creating person-specific skinning weights. While we limit our skinning weight to be sparse and symmetrical, the 9k skinning weight parameters still have too much freedom, which without sufficient regularization can result in artifacts. This motivates us to use a lower dimensional latent space to model the skinning weights generation. This approach is similar in spirit to GANFit \cite{GANFit} who optimizes the latent code of a GAN \cite{GAN} network.

\subsubsection{Skinning Weight Synthesis}
To train the skinning weight generator, we needed a set of ground-truth skinning weights. As our joint-based model generates meshes by joint transformation and skinning weight, if we perturb joint transformations, there should exist another set of skinning weights that result in about the same mesh. We generated 32k joint transformations by iteratively fitting joint transformations and skinning weights on scans. We saved these intermediate joint transformations and corresponding meshes during the fitting. We then froze the joint transformations and optimized skinning weights until they converged on each mesh. We end generating 32k sets of skinning weights ($w_s$).

\subsubsection{Network Design}
After generating the training data, we first follow COMA \cite{COMA} by building an autoencoder to encode a latent vector. We remove the encoder at test time and freeze the decoder, then fit the latent vector to generate skinning weights for each test scan. We found that the fitted latent vector during testing could be very different from training because the skinning weight training data are quite closed (because they are generated from only a small number of scans), thus the latent vector does not need to use all the space. We then remove the encoder and fine-tune the decoder with random noise ($Z$), similar to Generative Adversarial Network (GAN) to increase the coverage of $Z$ space. We found that pre-training the autoencoder is important as training only the decoder in GAN style is not stable. We choose fully connected (fc) layers with swish activation \cite{swish} as our network block. While graph convolution \cite{graph} is a compelling alternative, there is no obvious design for connecting skinning weights across different joints, so we let the network learn these connections.

The encoder $E$ reduces the dimension from 8,990 to 50 (Fig.\ref{fig:pipeline}), and the discriminator $C$ maps the dimension from 50 to 1. While the encoder and discriminator do not have to be compacted because we do not use them at test time, our decoder ($D$) maps $Z$ from 50 to 8,990, and is suffering the large number of parameters, especially on the late fc layers. Reducing the feature dimension of these fc layers causes the network to struggle to learn, so we keep the dimension large, but reuse the fc layer weight on different portions of the feature. To be specific, let $X$ denote the input dimension into fc layer, and $Y$ is the output dimension. Instead of learning $X*Y$ weights, we split $X$ into $n$ groups of $\frac{X}{n}$ inputs and only learn a single weight $\frac{X}{n}*\frac{Y}{n}$ to reuse on all the $n$ groups. This reduces the number of parameters by a factor of $n^2$. We call this: group-wise fully connected layer inspired by group-wise convolution from MobileNet \cite{mobilenet}. This can also be viewed as reshaping the 1D input $X$ into 2D shape of $n*\frac{X}{n}$, and apply 2D convolution with $\frac{Y}{n}$ filters of kernel size $1*\frac{X}{n}$.

\subsubsection{Losses}
We follow WGAN \cite{WGAN} where the discriminator loss on batch size ($m$) is:

\begin{equation}
L_C = \frac{1}{m}\sum_{i=1}^{m} (C(J_i) - C(D(Z_i)))
\end{equation}

We also add sparsity loss to the generator to prevent our neural skinning weights ($w_s$) from deviating too far from the learned linear skinning weights from section \ref{sec:skin}. We use the sparsity loss weight: $\lambda_0$ = 0.05 and adversarial loss weight: $\lambda_1$ = 0.1 throughout the experiments. Our generator loss is:

\begin{equation}
L_G = \sum_{i=1}^{m}(|J_i-D(Z_i)| + \lambda_0 |D(Z_i)| - \lambda_1 C(D(Z_i)))
\end{equation}

The generator and discriminator loss remain the same on both autoencoder and generator. Only the input changes from skinning weight ($w_s$) to noise ($Z$).

\subsection{Expression Modeling}
Because our joints have semantic meaning, we can use 12 of them to model a subset of facial expressions directly. However, to make sure our model has enough capacity to model all other expressions (such as cheek puff), we simply add 24 of the 47 blendshapes from \cite{MultiFaceRetarget19} as each of them represents one specific expression. The full list of blendshapes can be found in the supplemental material. To apply expression blendshapes on top of the neural skinning weight model, we fit joint transformation and neural skinning weight iteratively, then fit expression coefficients, and repeat the whole process for a few cycles. We use the loss function from Eq. \ref{loss} through out every stage.

\section{Expertimental Setup}
\subsection{Datasets}
\subsubsection{Retopologized scans:} We collected 94 scans, then manually retopologized them into our template. We then split 85 for training, 2 for validation and 7 for testing. These 85 scans were used to learn a global skinning weight in section \ref{sec:skin}. For neural skinning weights, we fit joint transformation with hand-painted skinning weights for 3k iterations, and saved these joint transformations on intermediate iterations. We did not use every iteration because if the iterations were too closed, the joint transformation had similar values. We then fit the global skinning weights across all the scans and refit joint transformations, then repeated the process for 5 cycles, resulting in 8k joint transformations. We sparsely perturbed them by 5\% for 3 more times to collect 32k joint transformations in total. We fit skinning weights on these joint transformations until converge on each of the corresponding retopologized scan. We split these 32k skinning weights into 29k training set, 680 validation set, and 2.4k test set.

\subsubsection{Separate testing scans} For external comparison, we evaluate the identity part of our model on BU-3DFE \cite{bu3dfe} 100 neutral scans, and evaluate the expressiveness of our model on 480 BU-3DFE scans (10 male and 10 female with all the scans except neutral). We setup a small rig with 1 Microsoft Azure Kinect (for depth) and 3 point grey cameras (for high-quality texture) and used it to collect, fit and visualize 70 scans to test the effectiveness of our model.

\subsection{Evaluation Metric}
We use point-to-point error on the retopologized test set as we already have the correspondences between predicted mesh and ground-truth scan. For BU-3DFE, we follow scan-to-mesh distance from FLAME. Unlike FLAME, our joint-based model automatically aligns with the root as part of fitting, so we do not need an extra alignment process.

\subsection{Training Configuration}
Our joint-based model was created by adding joints and painting skinning weights to a generic base mesh in Blender. We exported the model to .npz file and fit the parameters on Tensorflow. We use AdamOptimizer throughout the training. We fit joint transformations using a learning rate of $10^{-3}$ for 3k iterations. We fit the global skinning weights across the training set using a learning rate of $10^{-4}$ and batch size of 85 for 3k iterations. We fit joint transformations and skinning weights iteratively for 5 cycles. On a single Nvidia V100, fitting joint transformations took 1 minute per scan ($\sim$1.5 hours on training set), and fitting the skinning weights for 5 cycles took $\sim$20 hours. Neural skinning weights were trained using a learning rate of $10^{-3}$ during the first phase of training the autoencoder, and learning rate of $10^{-4}$ during the second phase of fine-tuning the decoder with a batch size of 85 for 30k iterations. It took $\sim$40 hours to train neural skinning weights on a V100 GPU. At test time, we built the correspondence on point clouds using NICP. We fit identity by fitting joint transformation for 500 iterations and $Z$ for 500 iterations for 2 cycles, then fit expressions for 500 iterations, then repeated the iterative fitting between identity and expressions for 2 cycles. The fitting took $\sim$2 minutes with a Nvidia 1080Ti.

\subsection{Neural Skinning Weight Architecture}
Our neural skinning weights consists of encoder, decoder and discriminator (Fig.\ref{fig:pipeline}). The basic building block is fc layer and swish activation. We use both fc layer weight and bias on the encoder and discriminator, but only use fc layer weight on the decoder. We remove fc layer bias from the decoder because if we initialize the input noise $Z$ as zero, our block will always output zero. We then designed the final skinning weights to be a summation between learned linear skinning weights from section \ref{sec:skin}, and the residual from our neural network. At test time, we first set $Z$ to zero and fit joint transformations to convergence, then fit $Z$ to generate a new skinning weight and repeat the cycle. Another reason for removing fc layer bias is to decrease the network size by half. 

As we do not use encoder and discriminator during the testing, they can be large. Each fc layer in our encoder reduces the dimensions by half from 8990 to 50. Our discriminator dimension goes from 50 to 256 in 4 layers, and another 4 layers to decrease the dimension to 1. The decoder contains 6 layers where the first 3 fc layers increase the dimension from 50 to 250, and the last 3 group-wise fc layers dimensions are 500 ($n$=2), 1100 ($n$=5) and 8990 ($n$=10), where $n$ is the number of groups.

\begin{figure}[!]
\centering
\includegraphics[width=1.0\linewidth]{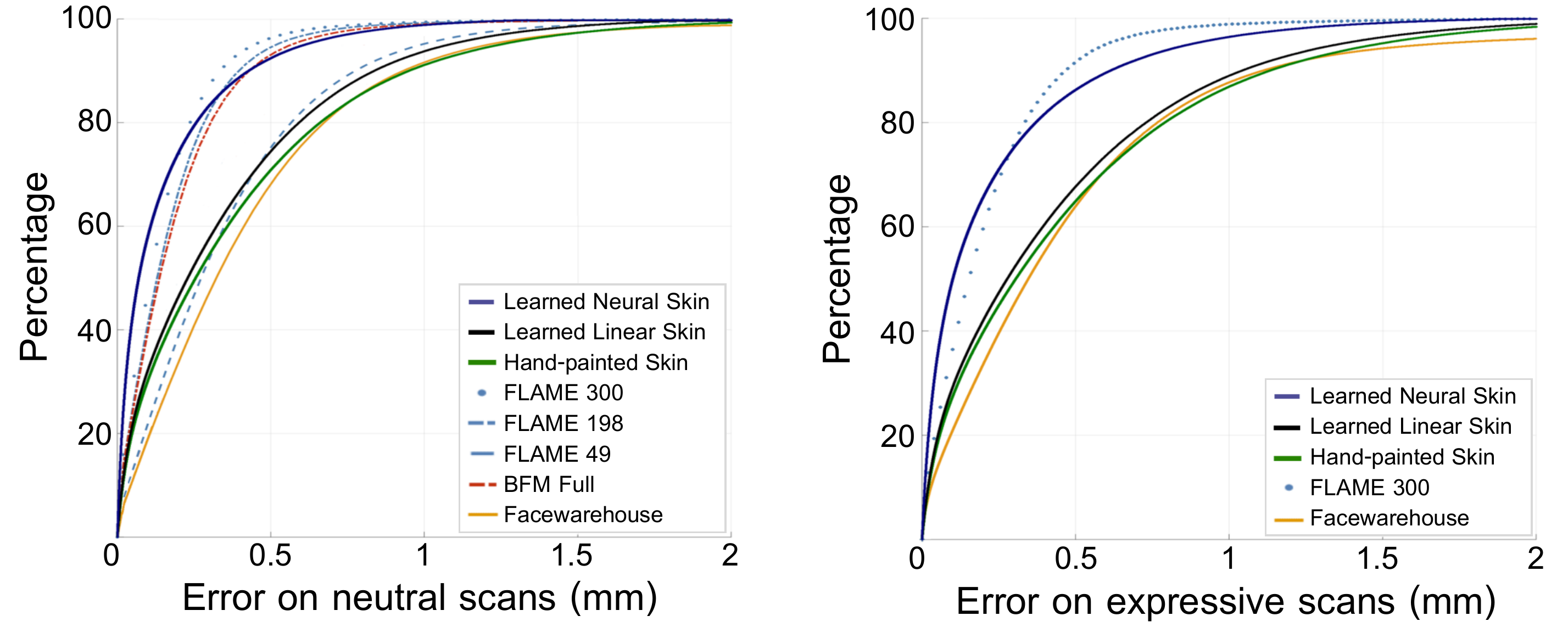}
\caption{External comparison on BU-3DFE. Our model is much more compact than the counterpart model with about the same scan-to-mesh error. The improvement of each skinning weight learning step also holds true.}
\label{fig:curves}
\end{figure}

\section{Results}
\subsection{Ablation Study}
\label{sec:ablation}
\noindent\textbf{Known Correspondence}
We conduct our experiments in a control environment on high quality scans with known correspondence to eliminate external factors. Table \ref{tab:retopo} shows that, in this control environment, hand-painted skinning weight already achieves a reasonable error (i.e. 0.41mm), while learning the linear skinning weight across training scans generalizes to the test set and further reduces the error. However, with our learned neural skinning, the error is reduced drastically from 0.41 to 0.11, which indicates that our neural skinning indeed increases joint-based model capacity very significantly. This is because each of the scan can have their personalized skinning weight, instead of just a single skinning weight that is shared across all the scans. This is a clear conclusion on high quality, known correspondence dataset. \\

\begin{table}[h]
\centering
\caption {Point-to-point error on known correspondence test set. Our learned skinning weights demonstrate improvement of model capacity on every step.}
\begin{tabular}{|l|c|}
\hline
\textbf{Model} & \textbf{Error (mm)} \\ \hline
(1) Hand-painted skinning   & 0.41 $\pm$ 0.02                \\ \hline
(2) Learned Linear skinning          & 0.34 $\pm$ 0.02                \\ \hline
(3) Learned Neural skinning          & 0.11 $\pm$ 0.01                \\  \hline
\end{tabular}
\label{tab:retopo}
\end{table}

\begin{figure}[!t]
\centering
\includegraphics[width=0.8\linewidth]{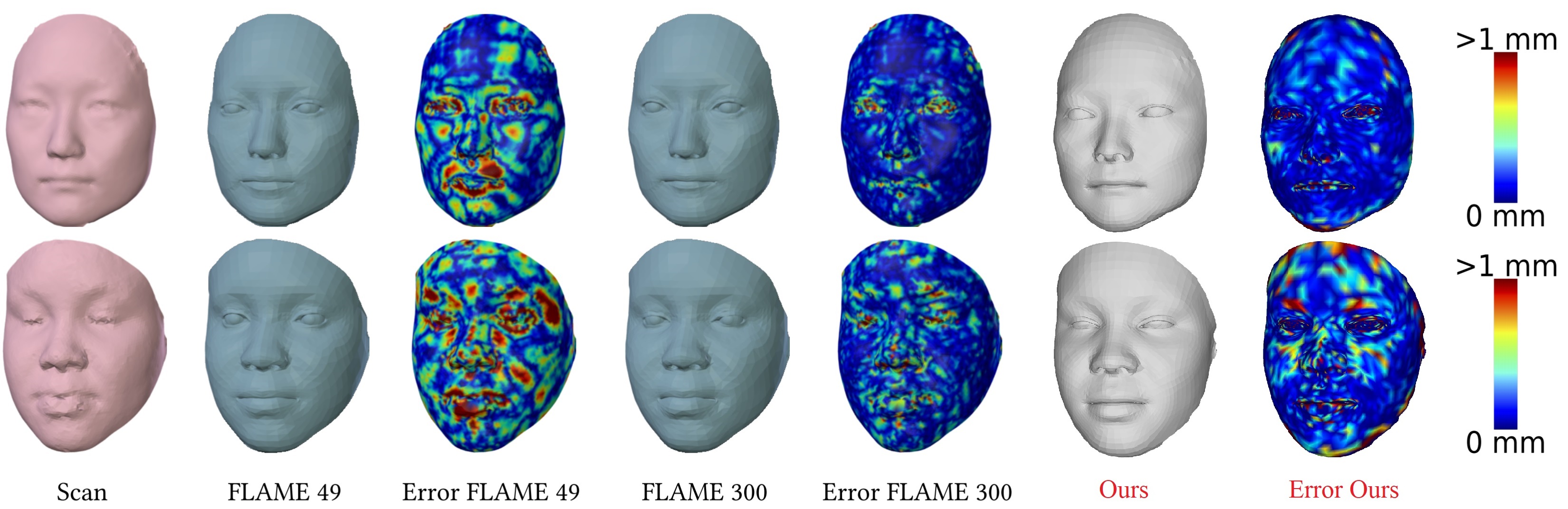}
\caption{Visual comparison between FLAME and neural skinning model on BU-3DFE.}
\label{fig:bu_vs_flame}
\end{figure}

\noindent\textbf{Unknown Correspondence} To compare to external methods, we study our model on BU-3DFE scans which are noisier than our training set. The correspondence is not known, so it could introduce correspondence error from ICP. Nevertheless, Fig.\ref{tab:bu3dfe_id},\ref{fig:curves} show that the improvement gap between our proposed skinning weights still hold true. For neutral face fitting, thanks to the novel joint based representation, even with human manual design and no training data, hand-painted skinning weight performs better than Facewarehouse \cite{Cao3d}. With linear skinning weight that learned from only 85 scans, we can achieve comparable performance with FLAME 49 while being 30x more compact (Table. \ref{tab:bu3dfe_id}), which is equivalent to only 1.6 PCA basis. With our neural skinning weight (solid blue curve), although the model size increases, it is still 20x more compact than FLAME 300 while having similar capacity (Fig.\ref{fig:bu_vs_flame}).

On expressive scans, the error increases across all the face model due to the non-rigid deformation of human face, and the increase of scan noise, especially on the surface inside the mouth of surprise scans. In spite of error increase, we observe similar improvement on our propose skinning weight (Table \ref{tab:bu3dfe_all} and Fig.\ref{fig:curves}). Our neural skinning model supports 6 expression in BU-3DFE out of the box and can achieve 0.239mm error. Adding expression blendshapes improves the error to be comparable with FLAME 300 while remaining 10x more compact. We also outperform Nonlinear 3DMM \cite{nonlinear3dmm} on BU-3DFE with Chamfer distance of 0.00083, 0.00075, 0.00077 for Nonlinear 3DMM, FLAME and ours respectively.

\begin{table}[h]
\centering
\caption{Model size and RMS error on BU-3DFE neutral scans. Our neural skinning model is comparable to FLAME 300 with 20x smaller model size. We only count the identity part of the model in \#Float.}
\begin{tabular}{|l|c|c|c|c|}
\hline
\textbf{Model}                           & \textbf{\#Float}  & \textbf{vs (3)} & \textbf{vs (1)} & \textbf{Error (mm)} \\ \hline
(1) FLAME 300 \cite{FLAME}               & 4.52M  & 0.16x    & 1x           &  0.158 $\pm$ 0.172          \\ \hline
(2) FLAME 49 \cite{FLAME}                & 738K   & 1x       & 6.1x         &  -                          \\ \hline
(3) Facewarehouse \cite{facewarehouse}   & 1.73M  & 0.43x    & 2.61         & 0.437 $\pm$ 0.474           \\ \hline
(4) BFM full \cite{BFM2009}              & 31.8M  & 0.02x    & 0.14x        &  -                          \\ \hline
(5) Hand-painted skinning                & 24.7K    & 29.9x    & 183x       & 0.375 $\pm$ 0.387           \\ \hline
(6) Learned linear skinning              & 24.7K    & 29.9x    & 183x       & 0.338 $\pm$ 0.346           \\ \hline
(7) Learned neural skinning              & 225.4K   & 3.3x     & 20.1x      & 0.153 $\pm$ 0.206           \\ \hline
\end{tabular}
\label{tab:bu3dfe_id}
\end{table}

\begin{table}[h]
\centering
\caption{Model size and RMS error on BU-3DFE expressive scans. Our neural skinning model is comparable to FLAME 300 with 10x smaller model size.}
\begin{tabular}{|l|c|c|c|}
\hline
\textbf{Model}                           & \textbf{\#Float}  & \textbf{vs (1)} & \textbf{Error (mm)} \\ \hline
(1) FLAME 300 \cite{FLAME}               & 6.03M     & 1x       & 0.211 $\pm$ 0.261           \\ \hline
(2) Facewarehouse \cite{facewarehouse}   & 79.42M    & 0.08x    & 0.558 $\pm$ 0.670           \\ \hline
(3) Hand-painted skinning                & 401.7K    & 15x    & 0.432 $\pm$ 0.518           \\ \hline
(4) Learned linear skinning              & 401.7K    & 15x    & 0.405 $\pm$ 0.494           \\ \hline
(5) Learned neural skinning              & 602.4K    & 10x    & 0.218 $\pm$ 0.289           \\ \hline

\end{tabular}
\label{tab:bu3dfe_all}
\end{table}

\subsection{Applications}
As we proof the effectiveness and compactness of our model in the section \ref{sec:ablation}, in this section, we show our results in real world applications.    \\

\noindent\textbf{Noisy Depth Fitting}
We demonstrate our result on consumer rig. We use depth from a single Microsoft Azure Kinect \cite{kinect} which suffers from multipass issue (cause the nose to shrink), missing parts due to IR absorbant material, and the depth itself is noisy. As our model can recover the full head from single depth map (the missing area will be automatically imply from the joint-based model prior), we use 3 point-grey cameras only for extracting full frontal texture. Fig.\ref{fig:Kinect} illustrates the effectiveness of our fitting result. Capturing the subject takes a split second and fitting takes 2 minutes on our setting.  \\

\noindent\textbf{RGB Image Fitting}
Like any previous face models, our model also supports 2D image based 3D face reconstruction. We conducted such an experiment on RGB images from Facewarehouse \cite{facewarehouse}. We use 2D landmarks as the only loss to fit the geometry. Fig.\ref{fig:2d} shows a few fitting results. As we can see, our model can fit reasonably well for even large facial expressions.

\subsection{Model Benefit}

\noindent\textbf{Editing \& Accessorizing}
As our model is designed with comprehensible list of joints, artist can tweak the joints to edit facial mesh. Adding accessory (e.g., hat, glass) is also possible, as most of the industrial computer graphic software (e.g., Blender, Maya) supports skinning weight transfer in just a few click. The artist can bind the new accessory to one of our joint and transfer skinning weight from the closest vertex. Changing hairstyle, attaching teeth, beard, whisker and mustache are also doable as well. The pipeline figure for adding accessories, and more demo video can be found in supplemental material.  \\

\noindent\textbf{Retargeting}
As our joints and expression blendshapes have semantic meaning, we can transfer the expression and pose into puppet (Fig. \ref{fig:2d}) or other face scans.

\section{Conclusion}
We propose a new 3D face representation by using joint-based face rig as the face model. We designed our model to be very compact, yet, preserve strong capacity by learning neural skinning weights. We evaluate our model on retopologized scans, BU-3DFE, visualize on Azure Kinect, and 2D images. Our model enjoy the benefit of facial mesh editing and accessorizing. As we can reasonably fit our face model to point cloud and 2D image, our future work will be learning a neural network to directly predict face model parameters \cite{Chaudhuri_2019_CVPR}, so that it could be possible to speed up our reconstruction for real-time applications.

\begin{figure*}[t]
\centering
\includegraphics[width=1.0\linewidth]{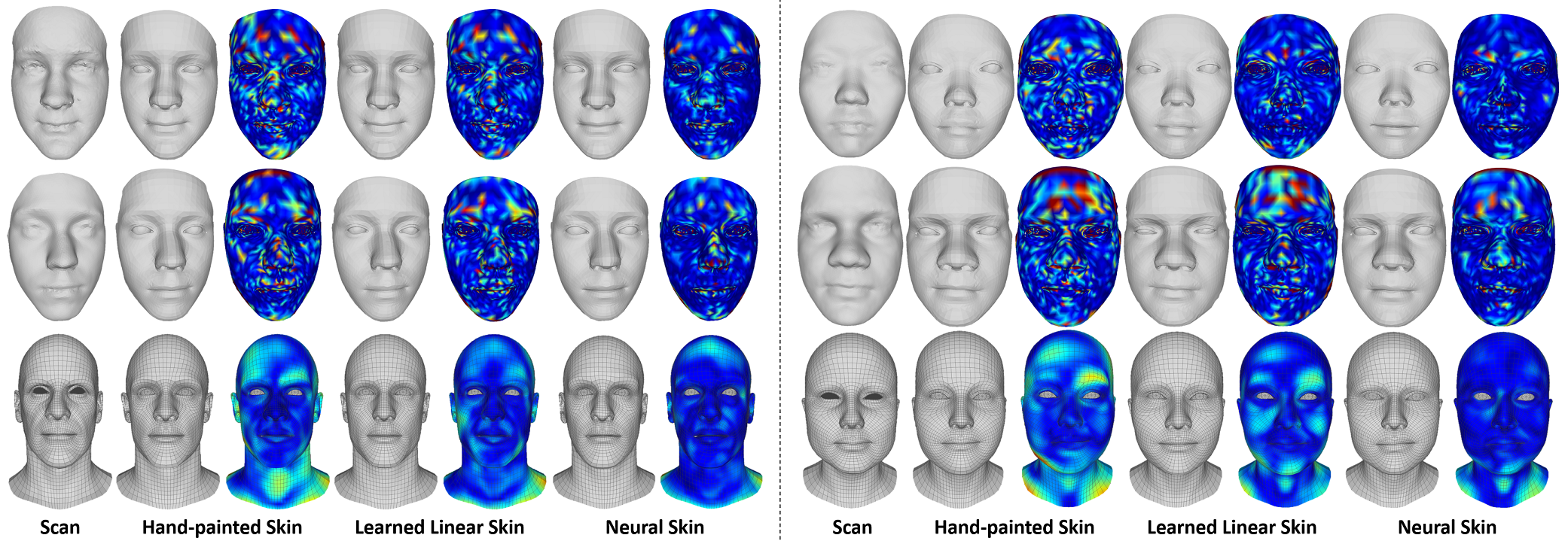}
\caption{Visualization of each step of improvement. Top 2 rows are BU-3DFE (top row scans are female), last row is our retopologized scan (right is female).}
\label{fig:more_results}
\end{figure*}

\begin{figure*}[t]
\centering
\includegraphics[width=1.0\linewidth]{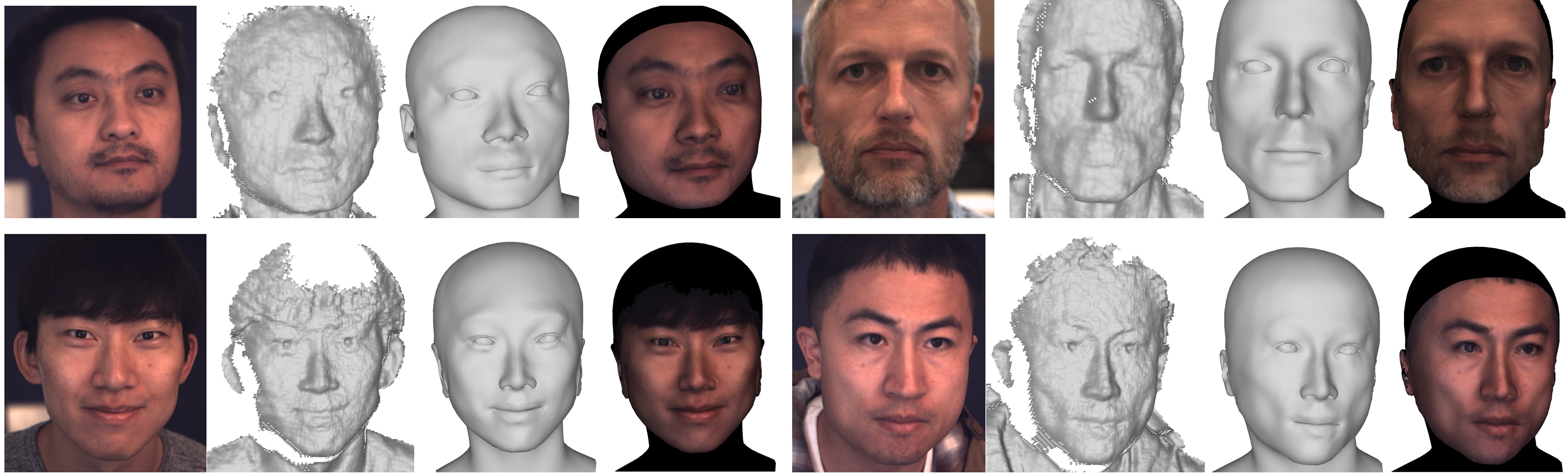}
\caption{Top: result on expressive scan from BU-3DFE (scan, result, error map). Bottom: result on Azure Kinect with point-grey cameras. Images are RGB (only for texture), depth (from Kinect), our fitted geometry, raycasted texture on our fitted geometry.}
\label{fig:Kinect}
\end{figure*}

\begin{figure*}[t]
\centering
\includegraphics[width=1.0\linewidth]{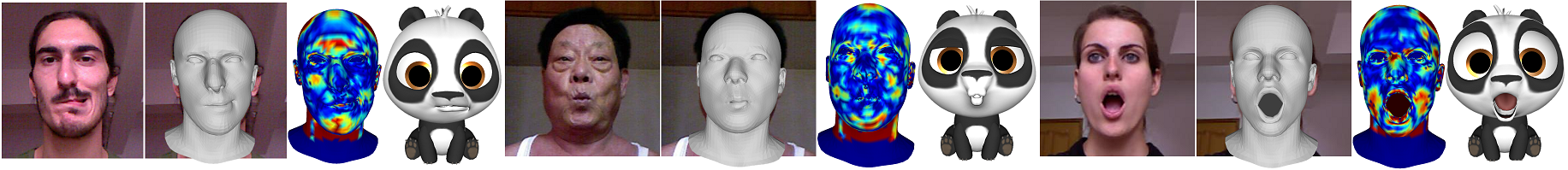}
\caption{3D face fitting from single 2D image by 2D landmark loss. Note that scan is only used for evaluation. The order is input, result, error map, and retargeting.}
\label{fig:2d}
\end{figure*}

\clearpage
\bibliographystyle{splncs04}

\begin{thebibliography}{10}
\providecommand{\url}[1]{\texttt{#1}}
\providecommand{\urlprefix}{URL }
\providecommand{\doi}[1]{https://doi.org/#1}

\bibitem{kinect}
Microsoft azure kinect,
  \url{https://azure.microsoft.com/en-us/services/kinect-dk/}

\bibitem{wrap3}
R3ds wrap 3, \url{https://www.russian3dscanner.com/}

\bibitem{zepeto}
Zepeto, \url{https://zepeto.me/}

\bibitem{BFM2009}
A 3D Face Model for Pose and Illumination Invariant Face Recognition (2009)

\bibitem{fernandez}
{Abrevaya}, V.F., {Wuhrer}, S., {Boyer}, E.: Multilinear autoencoder for 3d
  face model learning. In: 2018 IEEE Winter Conference on Applications of
  Computer Vision (WACV). pp.~1--9 (March 2018). \doi{10.1109/WACV.2018.00007}

\bibitem{NICP}
Amberg, B., Romdhani, S., Vetter, T.: Optimal step nonrigid icp algorithms for
  surface registration. In: CVPR. IEEE Computer Society (2007),
  \url{http://dblp.uni-trier.de/db/conf/cvpr/cvpr2007.html\#AmbergRV07}

\bibitem{WGAN}
Arjovsky, M., Chintala, S., Bottou, L.: {W}asserstein generative adversarial
  networks. In: Precup, D., Teh, Y.W. (eds.) Proceedings of the 34th
  International Conference on Machine Learning. Proceedings of Machine Learning
  Research, vol.~70, pp. 214--223. PMLR, International Convention Centre,
  Sydney, Australia (06--11 Aug 2017),
  \url{http://proceedings.mlr.press/v70/arjovsky17a.html}

\bibitem{bailey2018fast}
Bailey, S.W., Otte, D., Dilorenzo, P., O'Brien, J.F.: Fast and deep deformation
  approximations. ACM Transactions on Graphics (TOG)  \textbf{37}(4),  1--12
  (2018)

\bibitem{muscle}
Bao, M., Cong, M., Grabli, S., Fedkiw, R.: High-quality face capture using
  anatomical muscles. In: The IEEE Conference on Computer Vision and Pattern
  Recognition (CVPR) (June 2019)

\bibitem{3DMM}
Blanz, V., Vetter, T.: A morphable model for the synthesis of {3D} faces. In:
  Proceedings SIGGRAPH. pp. 187--194 (1999)

\bibitem{LSFM}
Booth, J., Roussos, A., Ponniah, A., Dunaway, D., Zafeiriou, S.: Large scale 3d
  morphable models. Int. J. Comput. Vision  \textbf{126}(2-4),  233--254 (Apr
  2018). \doi{10.1007/s11263-017-1009-7},
  \url{https://doi.org/10.1007/s11263-017-1009-7}

\bibitem{Cao3d}
Cao, C., Weng, Y., Lin, S., Zhou, K.: {3D} shape regression for real-time
  facial animation. ACM Transactions on Graphics  \textbf{32}(4) (Jul 2013)

\bibitem{facewarehouse}
Cao, C., Weng, Y., Zhou, S., Tong, Y., Zhou, K.: Facewarehouse: A 3d facial
  expression database for visual computing. IEEE Transactions on Visualization
  and Computer Graphics  \textbf{20}(3),  413--425 (March 2014)

\bibitem{MultiFaceRetarget19}
Chaudhuri, B., Vesdapunt, N., Wang, B.: Joint face detection and facial motion
  retargeting for multiple faces. In: IEEE Conference on Computer Vision and
  Pattern Recognition (CVPR) (2019)

\bibitem{Chaudhuri_2019_CVPR}
Chaudhuri, B., Vesdapunt, N., Wang, B.: Joint face detection and facial motion
  retargeting for multiple faces. In: The IEEE Conference on Computer Vision
  and Pattern Recognition (CVPR) (June 2019)

\bibitem{voxceleb2}
Chung, J.S., Nagrani, A., Zisserman, A.: Voxceleb2: Deep speaker recognition.
  In: INTERSPEECH (2018)

\bibitem{ear}
{Dai}, H., {Pears}, N., {Smith}, W.: A data-augmented 3d morphable model of the
  ear. In: 2018 13th IEEE International Conference on Automatic Face Gesture
  Recognition (FG 2018). pp. 404--408 (May 2018). \doi{10.1109/FG.2018.00065}

\bibitem{LYHM}
Dai, H., Pears, N., Smith, W.A.P., Duncan, C.: A 3d morphable model of
  craniofacial shape and texture variation. In: The IEEE International
  Conference on Computer Vision (ICCV) (Oct 2017)

\bibitem{Dai_2019_ICCV}
Dai, H., Shao, L.: Pointae: Point auto-encoder for 3d statistical shape and
  texture modelling. In: Proceedings of the IEEE/CVF International Conference
  on Computer Vision (ICCV) (October 2019)

\bibitem{JALI}
Edwards, P., Landreth, C., Fiume, E., Singh, K.: Jali: An animator-centric
  viseme model for expressive lip synchronization. ACM Trans. Graph.
  \textbf{35}(4),  127:1--127:11 (Jul 2016). \doi{10.1145/2897824.2925984},
  \url{http://doi.acm.org/10.1145/2897824.2925984}

\bibitem{Ferrari2015}
Ferrari, C., Lisanti, G., Berretti, S., Bimbo, A.D.: Dictionary learning based
  3d morphable model construction for face recognition with varying expression
  and pose. In: 2015 International Conference on 3D Vision. {IEEE} (Oct 2015).
  \doi{10.1109/3dv.2015.63}, \url{https://doi.org/10.1109/3dv.2015.63}

\bibitem{pablo_geomfromvideo}
Garrido, P., Valgaerts, L., Wu, C., Theobalt, C.: Reconstructing detailed
  dynamic face geometry from monocular video. In: {ACM} Trans. Graph.
  (Proceedings of SIGGRAPH Asia 2013). vol.~32, pp. 158:1--158:10 (November
  2013). \doi{10.1145/2508363.2508380},
  \url{http://doi.acm.org/10.1145/2508363.2508380}

\bibitem{GZCVVPT16}
Garrido, P., Zollh{\"o}fer, M., Casas, D., Valgaerts, L., Varanasi, K., Perez,
  P., Theobalt, C.: Reconstruction of personalized 3d face rigs from monocular
  video. {ACM} Trans. Graph. (Presented at SIGGRAPH 2016)  \textbf{35}(3),
  28:1--28:15 (2016)

\bibitem{GANFit}
Gecer, B., Ploumpis, S., Kotsia, I., Zafeiriou, S.: {GANFIT:} generative
  adversarial network fitting for high fidelity 3d face reconstruction. CVPR
  (2019), \url{http://arxiv.org/abs/1902.05978}

\bibitem{GAN}
Goodfellow, I., Pouget-Abadie, J., Mirza, M., Xu, B., Warde-Farley, D., Ozair,
  S., Courville, A., Bengio, Y.: Generative adversarial nets. In: Ghahramani,
  Z., Welling, M., Cortes, C., Lawrence, N.D., Weinberger, K.Q. (eds.) Advances
  in Neural Information Processing Systems 27, pp. 2672--2680. Curran
  Associates, Inc. (2014),
  \url{http://papers.nips.cc/paper/5423-generative-adversarial-nets.pdf}

\bibitem{mobilenet}
Howard, A.G., Zhu, M., Chen, B., Kalenichenko, D., Wang, W., Weyand, T.,
  Andreetto, M., Adam, H.: Mobilenets: Efficient convolutional neural networks
  for mobile vision applications (2017), \url{http://arxiv.org/abs/1704.04861},
  cite arxiv:1704.04861

\bibitem{mesoscopicgeometry}
Huynh, L., Chen, W., Saito, S., Xing, J., Nagano, K., Jones, A., Debevec, P.,
  Li, H.: Mesoscopic {Facial} {Geometry} {Inference} {Using} {Deep} {Neural}
  {Networks}. In: Proceedings of the IEEE Conference on Computer Vision and
  Pattern Recognition (CVPR). Salt Lake City, UT (2018)

\bibitem{Jiang_2019_CVPR}
Jiang, Z.H., Wu, Q., Chen, K., Zhang, J.: Disentangled representation learning
  for 3d face shape. In: Proceedings of the IEEE/CVF Conference on Computer
  Vision and Pattern Recognition (CVPR) (June 2019)

\bibitem{graph}
Kipf, T.N., Welling, M.: {Semi-Supervised Classification with Graph
  Convolutional Networks}. In: Proceedings of the 5th International Conference
  on Learning Representations. ICLR '17 (2017),
  \url{https://openreview.net/forum?id=SJU4ayYgl}

\bibitem{doi:10.1111/cgf.13830}
Li, K., Liu, J., Lai, Y.K., Yang, J.: Generating 3d faces using multi-column
  graph convolutional networks. Computer Graphics Forum  \textbf{38}(7),
  215--224 (2019). \doi{10.1111/cgf.13830}

\bibitem{FLAME}
Li, T., Bolkart, T., Black, M.J., Li, H., Romero, J.: Learning a model of
  facial shape and expression from {4D} scans. ACM Transactions on Graphics
  \textbf{36}(6),  194:1--194:17 (Nov 2017), two first authors contributed
  equally

\bibitem{liu2019neuroskinning}
Liu, L., Zheng, Y., Tang, D., Yuan, Y., Fan, C., Zhou, K.: Neuroskinning:
  Automatic skin binding for production characters with deep graph networks.
  ACM Transactions on Graphics (TOG)  \textbf{38}(4),  1--12 (2019)

\bibitem{Ankur2011}
Patel, A., Smith, W.: Simplification of 3d morphable models. In: Proceedings of
  the International Conference on Computer Vision. pp. 271--278 (2011).
  \doi{10.1109/ICCV.2011.6126252}, international Conference on Computer Vision
  ; Conference date: 06-11-2011 Through 13-11-2011

\bibitem{CFHM}
Ploumpis, S., Wang, H., Pears, N., Smith, W.A.P., Zafeiriou, S.: Combining 3d
  morphable models: A large scale face-and-head model. In: The IEEE Conference
  on Computer Vision and Pattern Recognition (CVPR) (June 2019)

\bibitem{swish}
Ramachandran, P., Zoph, B., Le, Q.V.: Searching for activation functions. ArXiv
   \textbf{abs/1710.05941} (2017)

\bibitem{ramach2017searching}
Ramachandran, P., Zoph, B., Le, Q.V.: Searching for activation functions (2017)

\bibitem{COMA}
Ranjan, A., Bolkart, T., Sanyal, S., Black, M.J.: Generating {3D} faces using
  convolutional mesh autoencoders. In: European Conference on Computer Vision
  (ECCV). vol. Lecture Notes in Computer Science, vol 11207, pp. 725--741.
  Springer, Cham (Sep 2018)

\bibitem{wildphotobasedreconstruction}
Roth, J., Tong, Y., Liu, X.: Adaptive {3D} face reconstruction from
  unconstrained photo collections. In: Proceedings of the IEEE Conference on
  Computer Vision and Pattern Recognition (CVPR) (2016)

\bibitem{300w}
Sagonas, C., Tzimiropoulos, G., Zafeiriou, S., Pantic, M.: 300 faces
  in-the-wild challenge: The first facial landmark localization challenge. In:
  IEEE International Conference on Computer Vision Workshops (ICCVW) (2013)

\bibitem{FML}
Tewari, A., Bernard, F., Garrido, P., Bharaj, G., Elgharib, M., Seidel, H.P.,
  Perez, P., Zollhofer, M., Theobalt, C.: Fml: Face model learning from videos.
  In: The IEEE Conference on Computer Vision and Pattern Recognition (CVPR)
  (June 2019)

\bibitem{selfsupervised_ayush}
Tewari, A., Zollh{\"o}fer, M., Garrido, P., Bernard, F., Kim, H., P{\'e}rez,
  P., Theobalt, C.: Self-supervised multi-level face model learning for
  monocular reconstruction at over 250 hz. In: Proceedings of the IEEE
  Conference on Computer Vision and Pattern Recognition (CVPR) (2018)

\bibitem{DBLP:conf/cvpr/TewariZ0BKPT18}
Tewari, A., Zollh{\"{o}}fer, M., Garrido, P., Bernard, F., Kim, H.,
  P{\'{e}}rez, P., Theobalt, C.: Self-supervised multi-level face model
  learning for monocular reconstruction at over 250 hz. In: 2018 {IEEE}
  Conference on Computer Vision and Pattern Recognition, {CVPR} 2018, Salt Lake
  City, UT, USA, June 18-22, 2018. pp. 2549--2559 (2018).
  \doi{10.1109/CVPR.2018.00270},
  \url{http://openaccess.thecvf.com/content\_cvpr\_2018/html/Tewari\_Self-Supervised\_Multi-Level\_Face\_CVPR\_2018\_paper.html}

\bibitem{nonlinear3dmm}
Tran, L., Liu, X.: Nonlinear {3D} face morphable model. In: Proceedings of the
  IEEE Conference on Computer Vision and Pattern Recognition (CVPR) (2018)

\bibitem{Weise:2011:RPF}
Weise, T., Bouaziz, S., Li, H., Pauly, M.: Realtime performance-based facial
  animation. In: ACM SIGGRAPH 2011 Papers. pp. 77:1--77:10. SIGGRAPH '11 (2011)

\bibitem{WU}
Wu, C., Bradley, D., Gross, M., Beeler, T.: An anatomically-constrained local
  deformation model for monocular face capture. ACM Trans. Graph.
  \textbf{35}(4),  115:1--115:12 (Jul 2016). \doi{10.1145/2897824.2925882},
  \url{http://doi.acm.org/10.1145/2897824.2925882}

\bibitem{incrementalfacetracking}
Wu, C., Shiratori, T., Sheikh, Y.: Deep incremental learning for efficient
  high-fidelity face tracking. ACM Trans. Graph.  \textbf{37}(6),
  234:1--234:12 (Dec 2018). \doi{10.1145/3272127.3275101},
  \url{http://doi.acm.org/10.1145/3272127.3275101}

\bibitem{bu3dfe}
Zhang, Z., Girard, J.M., Wu, Y., Zhang, X., Liu, P., Ciftci, U., Canavan, S.,
  Reale, M., Horowitz, A., Yang, H., Cohn, J.F., Ji, Q., Yin, L.: Multimodal
  spontaneous emotion corpus for human behavior analysis. In: The IEEE
  Conference on Computer Vision and Pattern Recognition (CVPR) (June 2016)

\end{thebibliography}



\section*{Supplementary material}

# JNR: Joint-based Neural Rig Representation for Compact 3D Face Modeling

Noranart Vesdapunt, Mitch Rundle, HsiangTao Wu, and Baoyuan Wang

Microsoft Cloud and AI
{`noves, mitchr, musclewu, baoyuanw`}@microsoft.com

(Supplemental Material)

Our joint-based neural rigging model (JNR) is a 3D face model designed by artist with learned neural skinning weight to increase model capacity while remains compact. We show additional figures and table in this supplementary to provide more visual results on both our retopologized scan and BU-3DFE, and full detail of our joint-based model to allow reproducibility. We also provide videos to demonstrate facial mesh editing and adding accessory.

**More Qualitative Results** We include more visual results of our retopologized scan in Fig.1 and more visual results of BU-3DFE in Fig.2. Both results are generated by neural skinning weight model. Similar to Fig.4-6 in the main submission.

**Joint-based Model Detail** We provide full detail of joint-based model in Table 1. Our joints are designed hierarchically to allow intuitive facial mesh editing and we list all the parent of each joint in Table 1. We also list all of our transformation parameters (rotation, translation, scale), transformation range, and a sample visualization of each joint transformation. Note that some joints do not have transformation parameter, but we still keep the joint to make editing more intuitive. For example, ear joint can move both left and right ear together. We list the expression blendshapes we used in Fig.3.

**Accessorizing Pipeline** To add accessory to our model, artist only need to register the accessory once to template model, and the skinning weight of accessory can be transferred from template model automatically by graphic software (e.g., Blender). Once the accessory is binded, any joint transformation from fitting result of our model can be used to transform accessory to attach to the fitted result. Pipeline diagram can be found in Fig.4.

**Demo Video** We attach demo videos for both facial mesh editing and adding accessory. Facial mesh editing video show the process of an artist who is changing facial feature by transforming correspondence joint. Accessorize video shows that once the accessory is attached by artist, no matter which identity or pose that our model transform into, the accessory is always attached.

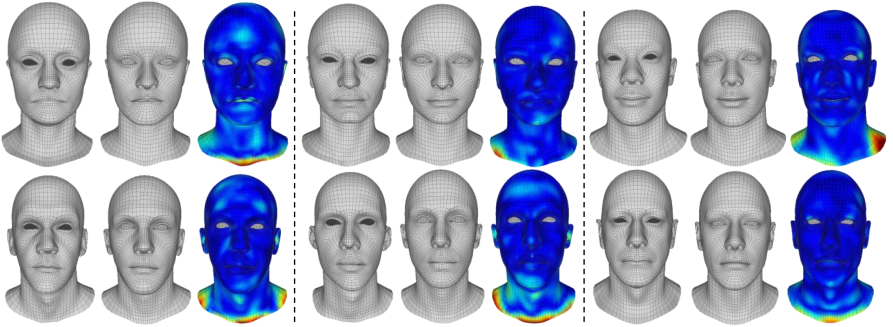

**Fig. 1**. Visualization of neural skinning weight model fitting result on retopologized scan test set. Images are ground-truth, fitted result, error map.

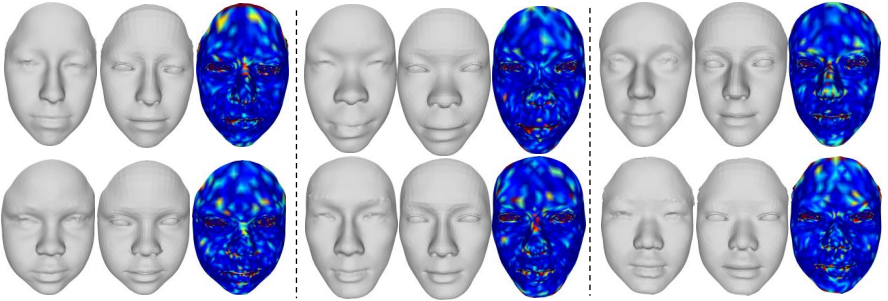

**Fig. 2**. Visualization of neural skinning weight model fitting result on BU-3DFE. Images are ground-truth, fitted result, error map.

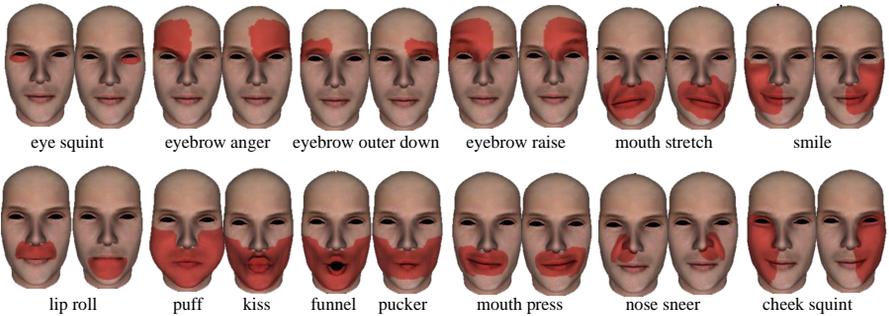

**Fig. 3**. Visualization of 24 expression blendshapes we used to expand the expression capacity. Red area highlights the vertex movement. Note that texture is only used for visualization purpose.

| Illustration | Name/Parent | Rotate | Translate | Scale | Sample |
|---|---|---|---|---|---|
| 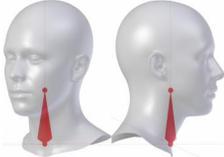 | Root /- | - | - | - | 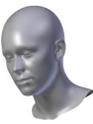 |
| 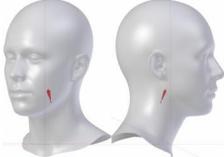 | Neck /Root | x: [-15, 15] y: [-2.4, 1] z: [-15, 15] | - | [0, 1.5] | 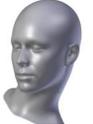 |
| 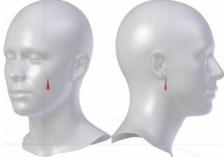 | Skull Root /Root | x: [-25, 25] y: [-15, 15] z: [-15, 15] | - | x: [0, 1.1] | 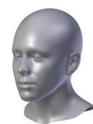 |
| 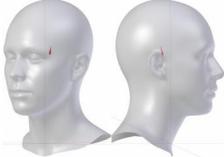 | Cranium /Skull Root | - | - | x: [-1, 1] z: [-1, 1] | 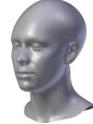 |
| 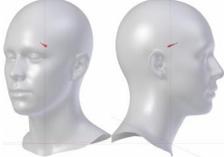 | Crown /Cranium | - | y: [-4, 4] | - | 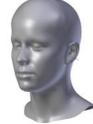 |
| 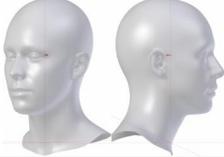 | Skull Center /Skull Root | - | - | - | 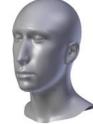 |
| 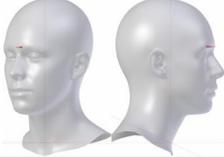 | Brow Center /Skull Center | - | z: [-1, 1] | y: [0, 1.5] | 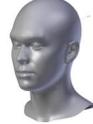 |

| Illustration | Name/Parent | Rotate | Translate | Scale | Sample |
|---|---|---|---|---|---|
| 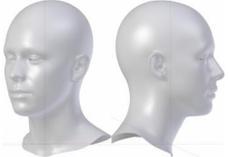 | Left Orbital /Skull Center | y: [-2,2] | x: [-1,1] z: [-0.3, 0.3] | - | 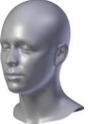 |
| 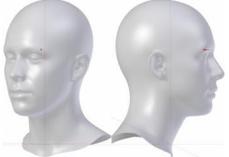 | Left Brow Outer /Left orbital | - | z: [-1, 1] | y: [0, 1.5] | 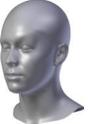 |
| 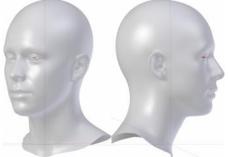 | Left Eye Socket /Left Orbital | y: [-6.5, 9] | x: [-1,1] y: [-1,1] | [0, 1.5] | 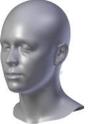 |
| 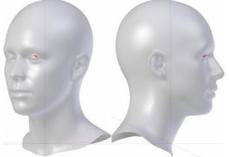 | Left Upper Eyelid /Left Eye Socket | x: [-6, 15] | - | - | 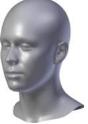 |
| 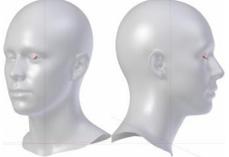 | Left Lower Eyelid /Left Eye Socket | x: [-11, 4] | - | - | 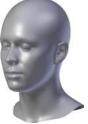 |
| 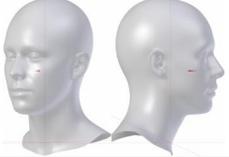 | Left Cheek Bone /Left Orbital | - | y: [-1.7, 2] | - | 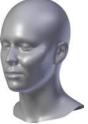 |
| 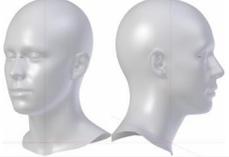 | Right Orbital /Skull Center | - | x: [-1,1] z: [-0.3, 0.3] | - | 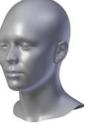 |

| Illustration | Name/Parent | Rotate | Translate | Scale | Sample |
|---|---|---|---|---|---|
| 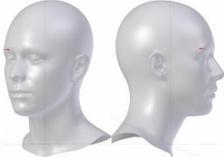 | Right Brow Outer /Right orbital | y: [-2,2] | z: [-1, 1] | y: [0, 1.5] | 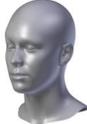 |
| 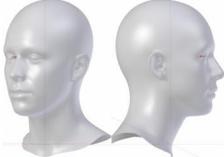 | Right Eye Socket /Right Orbital | y: [-6.5, 9] | x: [-1,1] y: [-1,1] | [0, 1.5] | 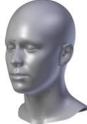 |
| 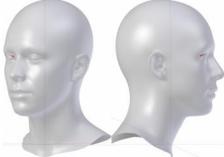 | Right Upper Eyelid /Right Eye Socket | x: [-6, 15] | - | - | 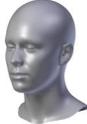 |
| 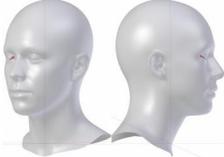 | Right Lower Eyelid /Right Eye Socket | x: [-11, 4] | - | - | 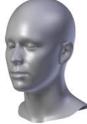 |
| 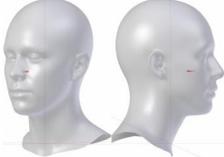 | Right Cheek Bone /Right Orbital | - | y: [-1.7, 2] | - | 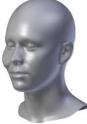 |
| 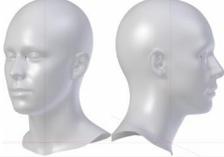 | Nose Root A /Skull Center | - | y: [-1, 1] z: [-1, 1] | x: [0, 1.2] | 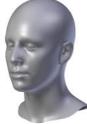 |
| 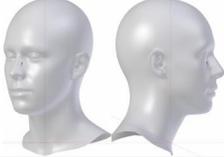 | Nose Root B /Nose Root A | - | - | - | 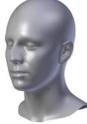 |

| Illustration | Name/Parent | Rotate | Translate | Scale | Sample |
|---|---|---|---|---|---|
| 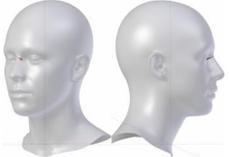 | Nose Bridge /Nose Root B | - | y: [-0.5, 0.8] | - | 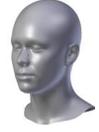 |
| 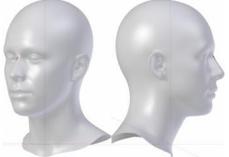 | Nose Ridge A /Nose Bridge | x: [-10, 8] z: [-2, 2] | - | y: [0, 1.5] | 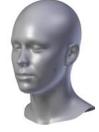 |
| 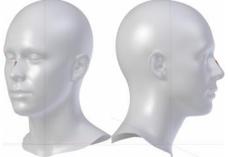 | Nose Ridge B /Nose Ridge A | x: [-2, 2] | - | y: [0, 1.5] | 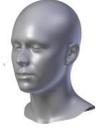 |
| 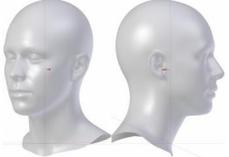 | Ears /Skull Root | - | - | - | 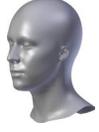 |
| 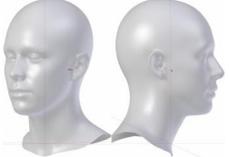 | Left Ear Position /Ears | - | x: [-1, 1] z: [-1.4, 1] | - | 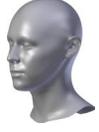 |
| 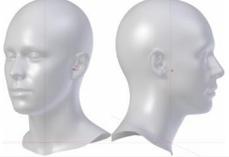 | Left Ear Rotate,Scale /Left Ear Position | z: [-30, 15] | - | [0, 1.3] | 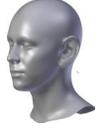 |
| 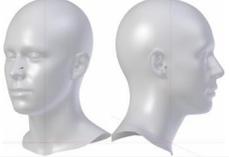 | Right Ear Position /Ears | - | x: [-1, 1] z: [-1.4, 1] | - | 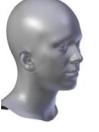 |

| Illustration | Name/Parent | Rotate | Translate | Scale | Sample |
|---|---|---|---|---|---|
| 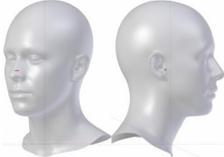 | Right Ear Rotate,Scale /Right Ear Position | z: [-30, 15] | - | [0, 1.3] | 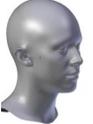 |
| 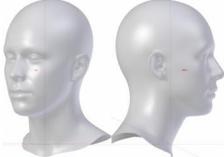 | Lower Face Root /Skull Root | - | - | - | 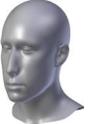 |
| 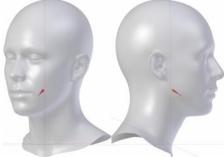 | Jaw Root /Lower Face Root | - | y: [-1.3, 0.6] z: [-1.3, 0.6] | x: [0, 1.1] | 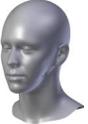 |
| 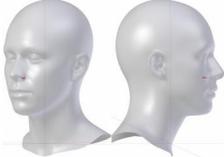 | Maxilla /Lower Face Root | - | y: [-0.5, 0.7] z: [-1, 1] | - | 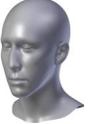 |
| 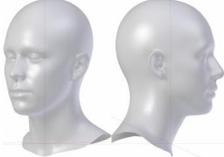 | Nose Base Position /Maxilla | - | x: [-0.4, 0.4] z:[-0.6, 0.6] | x: [0, 1.2] | 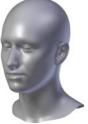 |
| 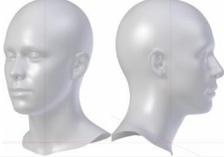 | Nose Base Rotate /Nose Base Position | x: [-8, 8] | - | - | 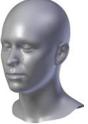 |
| 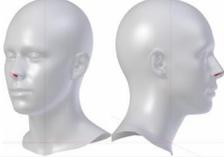 | Nose Tip /Nose Base Rotate | - | y: [-0.6, 0.6] z: [0.2, 1.1] | [0, 1.2] | 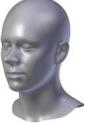 |

| Illustration | Name/Parent | Rotate | Translate | Scale | Sample |
|---|---|---|---|---|---|
| 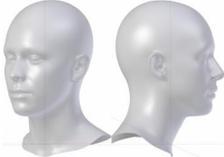 | Nose Nostril /Nose Base Rotate | x: [-8, 8] | - | [0, 1.2] | 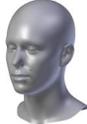 |
| 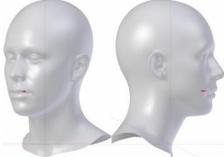 | Mouth Center Top /Maxilla | - | y: [-1.3, 0.6] z: [-1.3, 0.6] | x: [0, 1.2] | 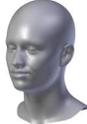 |
| 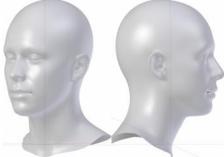 | Mouth Left Corner /Mouth Center Top | - | x: [-0.2, 0.2] y: [-0.2, 0.2] | - | 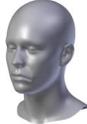 |
| 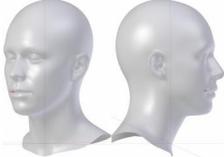 | Mouth Right Corner /Mouth Center Top | - | x: [-0.2, 0.2] y: [-0.2, 0.2] | - | 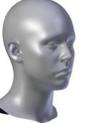 |
| 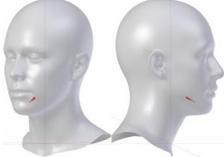 | Jaw Front /Mouth Center Top | z: [-0.9, 0.9] | y: [-0.8, 0.5] z: [-0.5, 0.5] | [0, 1.5] | 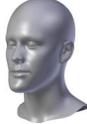 |
| 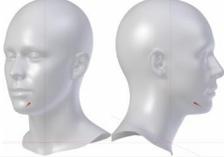 | Chin /Jaw Front | - | y: [-0.8, 0.5] z: [-0.5, 0.5] | [0, 1.5] | 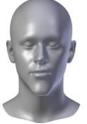 |
| 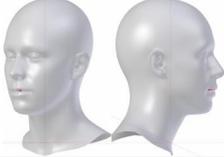 | Top Lip /Jaw Front | - | - | y: [0, 1.4] z: [0, 2] | 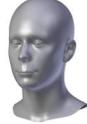 |

| Illustration | Name/Parent | Rotate | Translate | Scale | Sample |
|---|---|---|---|---|---|
| 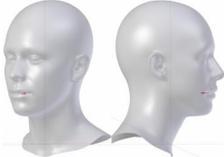 | Mouth Center Bottom /Mouth Center Top | - | y: [-0.3, 0.8] | - | 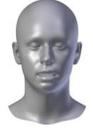 |
| 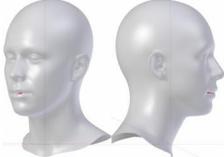 | Bottom Lip /Mouth Center Bottom | - | - | y: [0, 1.4] z: [0, 1.5] | 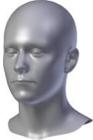 |
| 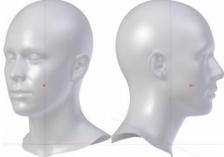 | Left Cheek /Maxilla | - | y: [-1.7, 2] | [0, 1.5] | 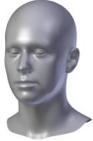 |
| 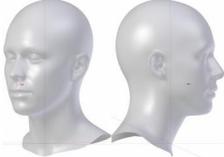 | Right Cheek /Maxilla | - | y: [-1.7, 2] | [0, 1.5] | 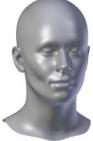 |
| 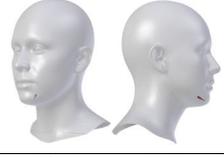 | Left Chin /Chin | - | x: [-0.2, 0.2] y: [-0.6, 0.6] | - | 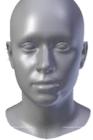 |
| 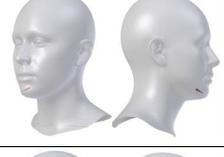 | Right Chin /Chin | - | x: [-0.2, 0.2] y: [-0.6, 0.6] | - | 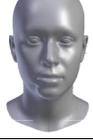 |
| 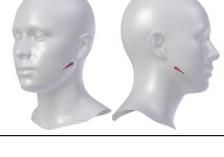 | Left Jaw Root /Jaw Front | - | y: [-0.8, 0.8] z: [-0.5, 1.5] | - | 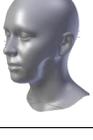 |

| 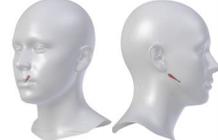 | Right Jaw Root /Jaw Front | - | y: [-0.8, 0.8]<br>z: [-0.5, 1.5] | - | 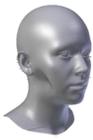 |
|---|---|---|---|---|---|
| 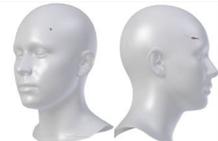 | Left Crow Outer /Crown | - | x: [-0.1, 0.1]<br>y: [-0.1, 0.1]<br>z: [-0.1, 0.1] | - | 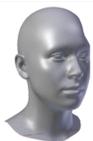 |
| 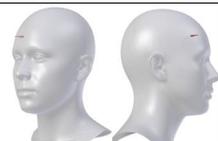 | Right Crown Outer /Crown | - | x: [-0.1, 0.1]<br>y: [-0.1, 0.1]<br>z: [-0.1, 0.1] | - | 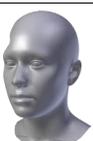 |

**Table 1**. Full detail of our joint-based model. First column is a visualization of joint (red) on template mesh on 2 views. Second column is the joint name and parent of each joint. Third to fifth column are transformation parameters with limit range. Note that rotation is in degree and scale without x,y,z annotation means all three axes share a single scale parameter. Last column is a visual example of the transformation from each joint.

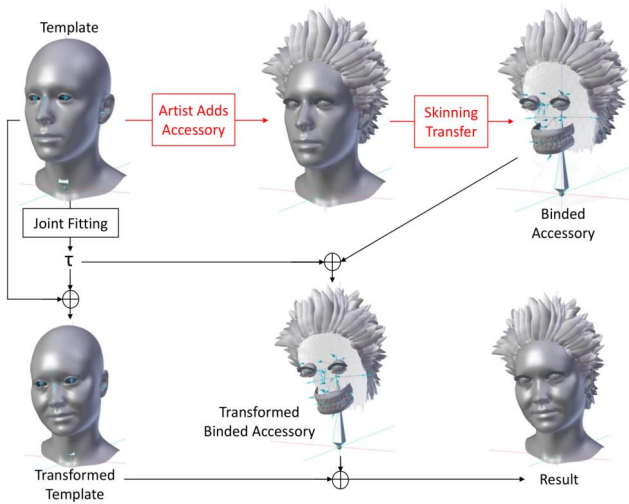

**Fig. 4**. Illustration of accessorizing. Artist only needs to add accessory to template once and transfers skinning weight to accessory. The binded accessory can be automatically applied by joint transformation ($\tau$) to any identity/pose.


\end{document}